\documentclass[runningheads]{llncs}
\usepackage{graphicx}
\usepackage{amsmath,amssymb} 
\usepackage{color}
\usepackage[width=122mm,left=12mm,paperwidth=146mm,height=193mm,top=12mm,paperheight=217mm]{geometry}

\usepackage{subfigure}
\usepackage[linesnumbered,ruled]{algorithm2e}

\def\ie{{\emph{i.e.}}}
\def\eg{{\emph{e.g.}}}
\def\etal{{\emph{et al.~}}}

\def\x{{\mathbf x}}
\def\X{{\mathbf X}}
\def\y{{\mathbf y}}
\def\c{{\mathbf c}}
\def\e{{\mathbf e}}
\def\z{{\mathbf z}}
\def\T{{\mathbf T}}
\def\t{{\mathbf t}}
\def\Y{{\mathbf Y}}
\def\E{{\mathbf E}}
\def\U{{\mathbf U}}
\def\S{{\mathbf \Sigma}}
\def\V{{\mathbf V}}

\begin{document}
\pagestyle{headings}
\mainmatter

\title{Tracking Completion} 

\titlerunning{Tracking Completion}

\authorrunning{Yao Sui, Guanghui Wang, Yafei Tang, Li Zhang}

\author{Yao Sui$^1$, Guanghui Wang$^{1,4}$, Yafei Tang$^2$, Li Zhang$^3$}


\institute{$^1$Dept. of EECS, University of Kansas, Lawrence, KS 66045, USA\\$^2$China Unicom Research Institute, Beijing 100032, China\\$^3$Dept. of EE, Tsinghua University, Beijing 100084, China\\$^4$National Laboratory of Pattern Recognition, Institute of Automation, CAS, China\\
\email{suiyao@gmail.com, ghwang@ku.edu, \\tangyf24@chinaunicom.cn, chinazhangli@tsinghua.edu.cn}
}

\maketitle

\begin{abstract}
  A fundamental component of modern trackers is an online learned tracking model, which is typically modeled either globally or locally. The two kinds of models perform differently in terms of effectiveness and robustness under different challenging situations. This work exploits the advantages of both models. A subspace model, from a global perspective, is learned from previously obtained targets via rank-minimization to address the tracking, and a pixel-level local observation is leveraged simultaneously, from a local point of view, to augment the subspace model. A matrix completion method is employed to integrate the two models. Unlike previous tracking methods, which locate the target among all fully observed target candidates, the proposed approach first estimates an expected target via the matrix completion through partially observed target candidates, and then, identifies the target according to the estimation accuracy with respect to the target candidates. Specifically, the tracking is formulated as a problem of target appearance estimation. Extensive experiments on various challenging video sequences verify the effectiveness of the proposed approach and demonstrate that the proposed tracker outperforms other popular state-of-the-art trackers.
\end{abstract}

\section{Introduction}
Visual tracking is an important topic in computer vision for its various applications, such as video analysis, robotics, and visual surveillance. In general, tracking models can be mainly classified into two categories: global and local. Global model exploits the overall information that varies in the entire target region. Local model treats the target as a series of small image patches to focus on the changes in each small region. It has been demonstrated that the global model is robust to some holistic appearance changes, like illumination variations and pose changes \cite{Ross2007,Mei2009,Babenko2011,Hare2011}. The local model, on the other hand, is intrinsically effective to the challenges, such as partial occlusions and local deformations \cite{Adam2006,Liu2011,Jia2012,Liu2015}. This is because only some of the local patches are influenced by the distractive objects (noise contaminated regions), while the rest are considered to be noise-free. To effectively deal with various appearance changes, a robust tracker is desired to be able to exploit the advantages of both global and local tracking models.

In this work, we propose  to leverage the effectiveness of the global method in capturing the overall information, and augment it with a local model to promote the accuracy and robustness of the tracker. The proposed tracking model integrates both the global and the local methods. Two efficient while effective methods, \ie, subspace learning and pixel-level local observations, are designed, and a matrix completion approach \cite{Candes2010} is employed to integrate the two models. The fundamental idea of our approach is to estimate the appearance of the target over the global subspace model and a number of local observations. As a result, the target is accurately located by means of the similarity between the estimation and the target candidates (regions of interest in the frame). Substantially different from previous tracking methods, which test each target candidate and then determine the best one as the target, the proposed approach works in a reverse way, \ie, predicts the expected target and then verifies it against each target candidate. To this end, the following two issues need to be addressed.

\subsection{Subspace Method}
Subspace method is a classical algorithm in visual tracking \cite{Ross2007,Kwon2010,Wang2012,Sui2015b}. Under this paradigm, the temporally obtained targets are assumed to reside in a low-dimensional subspace. For this reason, the current target can be accurately represented by the subspace learned from the previously obtained targets. It has been demonstrated that subspace method is effective to some challenges, such as pose changes and illumination variations \cite{Hager1996,Kriegmant1996}. However, this method is unstable in the presence of partial occlusions. The underlying assumption of subspace method, from a stochastic perspective, is that the representation errors obey the independent and identically distributed (\emph{i.i.d.}) Gaussian with small variances. In the case of partial occlusion, however, the representation errors actually follow the \emph{i.i.d.} Laplace or other heavy tailed distributions, because these errors may be extremely large but sparse. Consequently, a sparse (Laplace prior) additive error term is often used to compensate the instability of subspace model \cite{Wang2013,Wang2013a,Zhang2012b}.

Inspired by the previous success, we exploit the subspace structure among the previously obtained targets by using a $rank$-minimization method, instead of computing orthogonal basis vectors as used in previous methods. We stack these targets into column vectors respectively and then combine these columns into a sample matrix. Since these targets are assumed to reside in a low-dimensional subspace, this sample matrix tends to be of low-rank. Thus, we minimize the rank of this sample matrix to exploit the subspace structure. Although several tracking methods also involve low-rank matrix estimation \cite{Zhang2012b,Sui2015a}, their subspace assumptions are quite different from ours. Zhang \etal \cite{Zhang2012b} assume that the target candidates in each frame reside in a low-dimensional subspace and construct the subspace in the representation (transform) domain. Sui \etal \cite{Sui2015a} consider the obtained targets and the surrounding background regions reside in a mixture of several subspaces, and exploit these subspaces using a low-rank graph.

\subsection{Integration with Local Method}
Local tracking model is intrinsically robust to partial occlusions. Thus, it is reasonable to combine the subspace based tracking model with a local method to compensate the sensitivity of the subspace method in the case of occlusions. Previous local methods often transform the target region into a series of local image patches of small sizes with or without overlap. Different from those, our approach forces the local patches to shrink to the size of $1\times1$, leading to the \emph{local observations}, \ie, directly use a number of pixels\footnote{We only use \emph{a number of} pixels from the target region; otherwise, it is, to some extent, equivalent to global method.}. Note that the goals of our approach and previous local methods are essentially the same: intending to sufficiently leverage the noise-free pixels (patches) and avoid the corrupted pixels (patches). In contrast to patch based methods, our approach considers the corrupted pixels as the unobserved values and intends to estimate them over the exploited target subspace. Intuitively, the pixel-level method may lose the relationship among the neighboring pixels, \ie, correlations, which is well exploited in the patch-level strategy. Compared to the patches, however, the observed pixels are more flexible and much easier to be manipulated.

Matrix completion approach \cite{Candes2010}, by its nature, can be used to integrate the global target subspace model and the local observation method. The subspace model provides a prerequisite to ensure the success of the matrix completion, while the local observation method leads to a number of observed pixels to promote the accuracy of the matrix completion in the estimation of the unobserved (missing) pixels. In return, the matrix completion also implicitly maintains the subspace structure during the estimation. As demonstrated in our experiments, the estimation accuracy with respect to the target candidates, under the subspace assumption, is consistent with the similarity to the previously obtained targets, which is responsible to the target localization.

\subsection{Contributions}
The subspace model, from a global perspective, is learned via $rank$-minimization to address tracking, and the local observation approach, from a local point of view, is simultaneously leveraged to augment the subspace. The matrix completion is employed to integrate the two methods.
\begin{itemize}
  \item Unlike previous methods, which emphasize on analyzing all fully observed target candidates for target localization, the proposed approach leverages each partially observed target candidate to estimate the target with the learned subspace model via the matrix completion.
  \item The target is located according to the estimation accuracy of the matrix completion. It is shown that, under the subspace constraint, the estimation accuracy with respect to the target candidates is consistent with the similarity to the previously obtained targets. As a result, the proposed tracker performs much better than its counterparts.
\end{itemize}

\section{Related Work on Tracking}
Subspace learning is a conventional but effective method in visual tracking. Ross \etal \cite{Ross2007} utilized incremental subspace learning method to represent the target and locate the target in terms of representation accuracy. Wang and Lu \cite{Wang2012} proposed to use 2D principal component analysis method to construct a target subspace in original image domain. Sui \etal \cite{Sui2015b} proposed a group sparse subspace learning method to alleviate the influence of the distractive objects. Wang and Lu \cite{Wang2014} employed a segmentation-like method to improve the robustness of subspace learning against occlusions. Zhang \etal \cite{Zhang2012b} developed a low-rank and sparse representation to exploit the subspace structure among the candidates. Wang \etal \cite{Wang2013} assumed the targets follow a Gaussian distribution (subspace prior) and the occlusions followed a Laplace distribution (sparsity prior).

There are extensive literatures on local tracking. Adam \etal \cite{Adam2006} represented the target as histograms over a series local image patches. Liu \etal \cite{Liu2011} developed a local sparse representation to describe the target. Jia \etal \cite{Jia2012} designed an assignment pooling feature based on local sparse representation to improve the target description. Zhong \etal \cite{Zhong2012} utilized the local method to develop a collaborative target model. Kalal \etal \cite{Kalal2012} leveraged a local method to achieve a discriminative learning method for tracking. Sui and Zhang \cite{Sui2016} constructed a locally low-rank and sparse representation to address tracking.

Many impressive tracking results are also achieved by various approaches beyond subspace and local methods. Hare \etal \cite{Hare2011} employed the structured output support vector machine to address tracking. Gao \etal \cite{Gao2014} analyzed the likelihood of a candidate to be the target by using Gaussian process regression. Henriques \etal \cite{Henriques2015} proposed a robust tracker via correlation filters from kernelized ridge regression point of view, achieving impressive performance.

\section{\emph{Rank}-Minimization and Matrix Completion}
Recently, there has been a significant interest in $rank$-minimization. Some typical applications include matrix completion \cite{Candes2010}, robust principal component analysis \cite{Candes2011}, and low-rank representation \cite{Liu2010}. $rank$-minimization focuses on the problem
\begin{equation}
\label{eq:rm}
\min_\X rank\left(\X\right),~~
s.t.~\Y=f\left(\X\right),
\end{equation}
where $\Y$ denotes the observation matrix, $f\left(\X\right)$ is a restrict function with respect to the variable $\X$, and the $rank\left(\X\right)$ returns the rank of the matrix $\X$. The above minimization has been demonstrated to be a NP-hard problem. In practical applications, the convex conjugate $\left\|\cdot\right\|_*$, named as the $trace$-norm, is often used to approximate the $rank\left(\cdot\right)$ function. The $trace$-norm is defined as a sum of the singular values of the input matrix. Note that the significance of $rank$-minimization is partially attributed to its close relation to the subspace method. Specifically, Eq. \eqref{eq:rm} is equivalent to principal component analysis if the restrict function $f\left(\cdot\right)$ is an identical function, \ie,
\begin{equation}
\label{eq:rm_pca}
\min_\X rank\left(\X\right),~
s.t.~\left\|\Y-\X\right\|_F^2\le\varepsilon,
\end{equation}
where $\X$ is the reconstructed version of $\Y$ over the subspace, $\varepsilon>0$ is a very small number, and $\left\|\cdot\right\|_F$ denotes the $Frobenius$-norm. In fact, the matrix variable can be decomposed into $\X=\U\S\V^T$ via singular value decomposition (SVD), where $\U$ and $\V$ are orthogonal matrices, and $\S$ is a diagonal matrix composed by the singular values of $\X$. It is clear that the columns of $\U$ form the basis vectors of the learned subspace, and the columns of $\S\V^T$ are the subspace representations (\ie, the principal components). It is also evident that minimizing the rank of $\X$ is equivalent to making the diagonal elements of $\S$ as sparse as possible. In practice, $\X$ is reconstructed only from a few columns of $\U$, which correspond to the locations of the non-zeros of $\S$'s diagonal elements, so that the rank of $\X$ is minimized, leading to a subspace reconstruction. In this case, $rank$-minimization is directly related to the subspace method.

Matrix completion \cite{Candes2010} is one of the most popular applications of $rank$-minimization. It can accurately recover a matrix with missing entries, even if some entries are corrupted by noise. It is mathematically formulated as
\begin{equation}
\label{eq:mc}
\min_\X~\left\|\X\right\|_*,~
s.t.~\mathcal{P}_\Omega\left(\X\right)=\mathcal{P}_\Omega\left(\Y\right),
\end{equation}
where $\X$ is the recovered matrix, $\Y$ is the observation matrix, of which only the entries indexed by the set $\Omega$ can be observed, and $\mathcal{P}_\Omega\left(\X\right)$ is a projection function such that $\left[\mathcal{P}_\Omega\left(\X\right)\right]_{ij}=\X_{ij}$ for $\left(i,j\right)\in\Omega$ and zero otherwise. The goal of Eq. \eqref{eq:mc} is to estimate the missing entries (outside of $\Omega$) in terms of the observed entries (indexed by $\Omega$) of $\Y$. By minimizing Eq. \eqref{eq:mc}, the missing entries can be recovered. The theoretical analysis and recovering conditions can be found in \cite{Candes2010} and the references therein. Many algorithms have been developed to solve matrix completion, such as inexact augmented Lagrange multiplier (IALM) \cite{Lin2010}, and variational Bayesian inference \cite{Babacan2011a,Babacan2011}.

\section{The Proposed Approach}
\subsection{Problem Statement}
We describe the target region in each frame by using a motion state variable defined as
\begin{equation}
\z=\left\{x,y,s\right\},
\end{equation}
where $x$ and $y$ denote the 2D position of the target, and $s$ denotes the scale coefficient. According to the motion state variable, we can crop out the corresponding region from the frame image. The cropped region is resized to a predefined value and stacked into a column vector, which is named as the \emph{appearance observation}.

Our goal is to construct an estimator that can reliably predict an expected target appearance in each frame. Specifically, given the appearance observations $\y_1,\y_2,\dots,\y_{k-1}$ of previously obtained targets in the $k$-th frame, we can estimate the target appearance in the current frame as
\begin{equation}
\hat{\y}_k=\varphi\left(\y_1,\y_2,\dots,\y_{k-1}\right)
\end{equation}
by using the estimator $\varphi\left(\cdot\right)$. To make the estimator as accurate as possible, some prior knowledge about the target appearance, is encouraged. Thus, the estimated appearance of the current target is reformulated as
\begin{equation}
\hat{\y}_k=\varphi\left(\y_1,\y_2,\dots,\y_{k-1}|\Phi\right)
\end{equation}
by incorporating the prior information $\Phi$. Then, we find a region, of which the corresponding appearance is most similar to $\hat{\y}_k$, as the target region in the current frame. Mathematically, given a set of the appearance observations $\mathcal{C}$ of all target candidates, the current target is located by
\begin{equation}
\label{eq:tl}
\y_k=\arg\min_{\c\in\mathcal{C}}\left\|\hat{\y}_k-\c\right\|.
\end{equation}

From a global perspective, the previously obtained targets are considered to reside in a low-dimensional subspace due to their high similarity in appearances. From a local point of view, local observations (partial target information) can be obtained to help the estimator make a more accurate prediction. As a result, the problem is solved by integrating both the global and the local information. We exploit the global correlation to handle the previously obtained targets, and leverage the local information to deal with the target priors.

\subsection{Estimator Design}
As presented above, the estimator is built on the two kinds of information: the appearance observations of previously obtained targets and the prior knowledge of the target. The designed estimator will be discussed below.

\emph{Target summarization}. In order to increase the computational efficiency, the tracking model employs a compact form, instead of using all appearance observations, to represent the previously obtained targets. Meanwhile, such a compact representation is also explored to maintain the subspace assumption. To this end, only a limited number of previously obtained targets, which can best describe the appearance changes of all the obtained targets, are employed as the estimation evidence of the estimator. We refer to the target template method \cite{Mei2009} to implement the target summarization, which is called \emph{target templates} hereafter.

\emph{Target priors}. In the proposed model, the prior knowledge is extracted directly from a number of pixels in the target region, because such direct partial observations are the best and the strongest prior information for the target. There is, however, an obvious paradox, \ie, we intend to estimate the target appearance, while the estimator needs to partially observe the target appearance first. For this reason, we observe a number of pixels from each target candidate, and the target candidate is employed to eliminate the paradox. Under the low-dimensional subspace assumption, the target is expected to be estimated accurately by the estimator among all target candidates. The underlying reason is straightforward: since the previously obtained targets span a low-dimensional subspace, while the current target can be well represented by this subspace.

Based on the preceding analysis, the matrix completion approach is a desirable estimator for our tracking model. On one hand, matrix completion is a reliable estimator to predict unobserved entries. On the other hand, it can implicitly maintain the subspace constraint through the $rank$-minimization.

Given an appearance observation, denoted by $\c$, of a target candidate in each frame\footnote{For the presentation simplicity, we use the term \emph{candidate} to stand for the appearance observation of the target candidate hereafter.}, we use a set $\Omega$ to index the observed pixels, and consider the rest as missing values. We first generate an \emph{observed candidate} $\c'$ by setting the pixels of $\c$ outside $\Omega$ to zeros and leaving the rest unchanged. Let a matrix $\T=\left[\t_1,\t_2,\dots,\t_n\right]$ denote the $n$ target templates, which are summarized from $\left\{\y_1,\dots,\y_{k-1}\right\}$. We construct a new matrix $\Y=\left[\T,\c'\right]$ and estimate the pixels outside $\Omega$ using matrix completion over $\Y$. For convenience, we use an equivalent form of Eq. \eqref{eq:mc} to address the matrix completion by introducing a slack variable $\E$.
\begin{equation}
\label{eq:mc_1}
\min_\X~\left\|\X\right\|_*,~
s.t.~\Y=\X+\E,~\mathcal{P}_\Omega\left(\E\right)=0.
\end{equation}
The above minimization problem \eqref{eq:mc_1} can be solved by the IALM approach \cite{Lin2010}. Let $\X^*=\left[\T^*,\x\right]$ denote the solution of Eq. \eqref{eq:mc_1}, where $\x$ is the estimated candidate over the observed candidate $\c'$.

\subsection{Target Localization}
Within the Bayesian sequential inference framework \cite{Isard1998,Arulampalam2002}, given all the obtained targets $\y_{1:k-1}$ in the $k$-th frame, the motion state of the $k$-th target, denoted by $\z_k$, is predicted by maximizing the posterior
\begin{equation}
\label{eq:MAP}
p\left(\z_k|\y_{1:k-1}\right)=\int p\left(\z_k|\z_{k-1}\right)p\left(\z_{k-1}|\y_{1:k-1}\right)d\z_{k-1},
\end{equation}
where $p\left(\z_k|\z_{k-1}\right)$ denotes the \emph{motion model}. Then, a target candidate is generated according to its motion state $\z_k$. Thus, the corresponding appearance observation, denoted by $\c$, is obtained and the posterior is updated by
\begin{equation}
\label{eq:updated_MAP}
p\left(\z_k|\c,\y_{1:k-1}\right)\propto p\left(\c|\z_k\right)p\left(\z_k|\y_{1:k-1}\right),
\end{equation}
where $p\left(\c|\z_k\right)$ denotes the \emph{observation model}. The target on the $k$-th frame, denoted by $\y_k$, is found by
\begin{equation}
\label{eq:target_location}
\y_k=\arg\max_{\c\in\mathcal{C}}p\left(\z_k|\c,\y_{1:k-1}\right),
\end{equation}
where $\mathcal{C}$ denotes the set of all the candidates that correspond to a series of regions sampled randomly in the frame according to the possibility $p\left(\c|\z_k\right)$.

The motion model in our work is defined as a Gaussian distribution $p\left(\z_k|\z_{k-1}\right)\sim\mathcal{N}\left(\z_k|\z_{k-1},\mathbf{\Sigma}\right)$, where the covariance $\mathbf{\Sigma}$ is a diagonal matrix, denoting the variances of 2D translation and scaling, respectively. We set $\mathbf{\Sigma}=diag\left\{3,3,0.005\right\}$ in our experiments. The observation model $p\left(\c|\z_k\right)$ reflects the likelihood of the candidate $\c$ to be the target.
As discussed above, a good candidate can be estimated accurately by the matrix completion under our subspace assumption. The accuracy is measure by means of the estimation errors. Let us define the observation model for a candidate $\c$ with the motion state $\z_k$ as
\begin{equation}
\label{eq:est_err}
p\left(\c|\z_k\right)\propto\exp\left(-\left\|\c-\x\right\|\right).
\end{equation}
For all the candidates and their corresponding motion states, the target in the $k$-th frame can be located using Eq. \eqref{eq:target_location}. Note that under the definition of the observation model, Eq. \eqref{eq:target_location} is equivalent to Eq. \eqref{eq:tl}, and yields the same result in the target location. The implementation details of the tracking algorithm is outlined in Algorithm \ref{alg:tracking}.
\begin{algorithm}[t]
\caption{Tracking Algorithm}
\label{alg:tracking}
{\scriptsize
\KwIn{index set $\Omega$ and target templates $\T$.}
\KwOut{the target located in the $k$-th frame.}
\For{each candidate $\c\in\mathcal{C}$}
{
Generate the observed candidate $\c'$ by setting $\c$'s entries outside $\Omega$ to zeros and leaving the rest unchanged. \\
Construct the matrix $\Y=\left[\T,\c'\right]$. \\
Obtain the estimated candidate $\x$ from Eq. \eqref{eq:mc_1}. \\
Compute the observation model from Eq. \eqref{eq:est_err}. \\
}
Locate the target from Eq. \eqref{eq:target_location}. \\
Update the index set $\Omega$ and the target templates $\T$. \\
}
\end{algorithm}

\begin{figure}[t]
  \centering
  \subfigure[]{
  \includegraphics[width=0.12\linewidth]{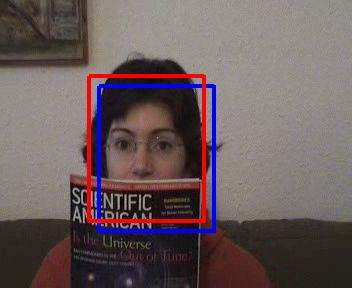}
  \label{fig:img}
  }
  \hfil
  \subfigure[]{
  \includegraphics[width=0.33\linewidth]{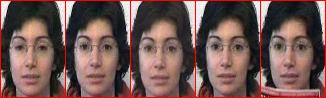}
  \label{fig:pre_objs}
  }
  \caption{(a) One frame image, where a good and a bad target candidate are marked in red and blue, respectively. (b) The representative target templates.}
  \label{fig:exam}
\end{figure}

\begin{figure}[t]
  \centering
  \subfigure[]{
  \includegraphics[width=0.1\linewidth,height=0.6in]{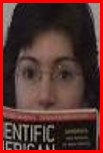}
  \label{fig:orgnl_obj}
  }
  \subfigure[]{
  \includegraphics[width=0.1\linewidth,height=0.6in]{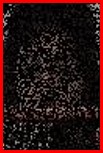}
  \label{fig:obsv_obj}
  }
  \subfigure[]{
  \includegraphics[width=0.1\linewidth,height=0.6in]{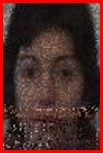}
  \label{fig:est_obj}
  }
  \subfigure[]{
  \includegraphics[width=0.1\linewidth,height=0.6in]{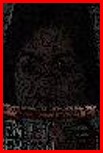}
  \label{fig:err_obj}
  }
  \subfigure[]{
  \includegraphics[width=0.1\linewidth,height=0.6in]{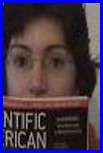}
  \label{fig:orgnl_trans}
  }
  \subfigure[]{
  \includegraphics[width=0.1\linewidth,height=0.6in]{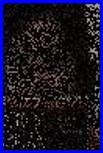}
  \label{fig:obsv_trans}
  }
  \subfigure[]{
  \includegraphics[width=0.1\linewidth,height=0.6in]{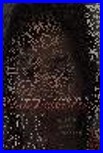}
  \label{fig:est_trans}
  }
  \subfigure[]{
  \includegraphics[width=0.1\linewidth,height=0.6in]{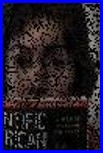}
  \label{fig:err_trans}
  }
  \caption{(a)-(d) The good candidate, its observed pixels, estimated result, and estimation error. (e)-(h) The corresponding results as (a)-(d) with respect to the bad candidate.}
  \label{fig:mc}
\end{figure}

Below is a demonstration of the proposed approach. As shown in Fig. \ref{fig:img}, two candidates are marked in red and blue, respectively. The representative target templates are shown in Fig. \ref{fig:pre_objs}. We crop out the two target candidates from the image and resize them to the same size as the target templates, as shown in Fig. \ref{fig:orgnl_obj} and \ref{fig:orgnl_trans}. Then, we sample a number of pixels of the two candidates at the same locations and use these pixels as the observed values, while the rest are treated as missing values, as shown in Figs. \ref{fig:obsv_obj} and \ref{fig:obsv_trans}, where the missing values are set to zeros. Next, we estimate the missing values of each candidate from Eq. \eqref{eq:mc_1}. Figs. \ref{fig:est_obj} and \ref{fig:est_trans} show the two estimated candidates, respectively. Their estimation errors are shown in Figs. \ref{fig:err_obj} and \ref{fig:err_trans}, respectively.
\begin{figure}[t]
  \centering
  \includegraphics[width=0.5\linewidth]{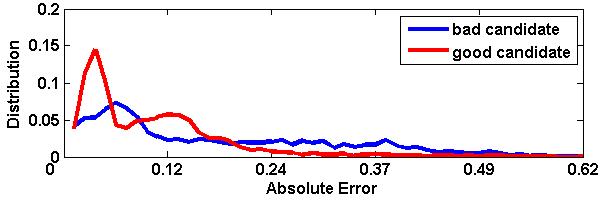}
  \caption{Distributions of the absolute estimation errors at the missing entries of the two candidates.}
  \label{fig:err_dist}
\end{figure}

From the above results, it is evident that the good candidate (in red) is estimated much more accurately than the bad one (in blue). As shown in Fig. \ref{fig:est_obj}, the estimated good candidate is rarely influenced by the distractive object (the magazine), however, the estimated bad candidate, as shown in Fig. \ref{fig:est_trans}, is quite different from its original version shown in Fig. \ref{fig:orgnl_trans}. Similar results can also be observed from their estimation errors. Most errors of the good candidate are small, and large errors only appear at the location of the distractive object, as shown in Fig. \ref{fig:err_obj}. In contrast, most errors of the bad candidate are large, and scatter all over the entire image, as shown in Fig. \ref{fig:err_trans}. Quantitatively, we also plot the distributions of the absolute estimation errors at the missing entries of the two candidates, as shown in Fig. \ref{fig:err_dist}. It can be seen that for most missing entries, the errors of the good candidates are much smaller than those of the bad ones. In addition, the residual errors of the good candidates normally converge faster than those of the bad ones because the good candidates better match the implicitly learned subspace via the $rank$-minimization. Typically, the matrix completion runs less than 30 iterations for good candidates, while about 40 iterations are required for bad candidates.

The good performance of the matrix completion in this case is attributed to two aspects: the low-dimensional subspace assumption on the previously obtained targets, and the local observations from the candidates. From a global point of view, the previously obtained targets span a low-dimensional subspace, which makes better representations of the good candidates, such that they can be estimated more accurately than the bad ones. From a local perspective, the local observations work as strong priors and promote the accuracy of the estimation. Since the index set $\Omega$ is determined according to the previously obtained targets, some pixels observed from the bad candidate may be located on the distractive object, leading to a more inaccurate estimation.

\subsection{Online Update}
During tracking, the appearance of the target varies on successive frames. Thus, we need to update the tracker automatically to accommodate these appearance changes. In each frame, a number of pixels of the candidates are sampled so as to alleviate the influence of the distractive objects. Therefore, the set $\Omega$ is updated for every frame to exclude those unexpected pixels. Meanwhile, the target templates $\T$ are updated accordingly, in order to accurately reflect these appearance changes and satisfy the constraint of low-dimensional subspace.

In our work, each pixel of an obtained target is associated to a weight that reflects the possibility of this pixel to be observed in the next frame. Initially, we set all these weights equally. As analyzed in the above demonstration shown in Figs. \ref{fig:exam}-\ref{fig:err_dist}, the estimation errors are normally large in the regions of the distractive objects (see Fig. \ref{fig:err_obj}). Thus, we adjust the weights in each frame to be inversely proportional to the corresponding estimation errors. To avoid that the observed pixels (they always have zero estimation errors) dominate the update, their weights are constrained during the computation. Finally, we draw the same number of entries randomly according to their weights and use these entries as the new index set $\Omega$.

Specifically, in the $k$-th frame, the weight of the the $j$-th pixel, denoted by $w_j^k$, is updated by
\begin{equation}
\label{eq:update_w}
w_j^k\propto
\begin{cases}
\begin{array}{ll}
\frac{1}{e_j^k}, & j\notin\Omega \\
\frac{1}{e_a+e_j^{k-1}\left(e_b-e_a\right)}, & j\in\Omega,
\end{array}
\end{cases}
\end{equation}
where $e_j^k$ denotes the estimation error of the $k$-th target in the $j$-th pixel, and $e_a$ and $e_b$ are determined by
\begin{equation}
e_{i_1}<e_{i_2}<\dots<e_a<e_m<e_b<\dots<\e_{i_N},
\end{equation}
where $e_m$ denotes the median value of the $N$ estimation errors, and $i_k\in\left\{1,2,\dots,N\right\}$. In the above equations, we divide the pixels into two categories and update their associated weights respectively. One category contains the pixels outside the index set $\Omega$, \ie, in the case of $j\notin\Omega$ for the $j$-th pixel of the $k$-th target. Among these pixels, the pixels with large estimation errors are unexpected to be observed in the next frame, since they have high possibilities to be located on the distractive objects. Thus, we directly set their associated weights inversely proportional to their estimation error $e_j^k$. The other category contains the pixels indexed by $\Omega$. Because these pixels are the observed ones in the current frame, \ie, they have zero estimation errors, they are expected to be observed in the next frame. In addition, in order to avoid that these pixels dominate the update, we deliberately decrease their possibilities to be observed to some extent. For this reason, we constraint the possibilities of these pixels within an appropriate range, or equivalently assign them certain errors within an range $\left[e_a,e_b\right]$. In practice, the median of the target estimation errors in last frame, \ie, the $\left(k-1\right)$-th target, is a reasonable reference in setting $e_a$ and $e_b$, such that their values are not being set too low or too high. In our experiments, $e_a$ and $e_b$ are set to the errors just below and above the median error, respectively.

\begin{figure}[t]
  \centering
  \subfigure[]{
  \includegraphics[width=0.1\linewidth,height=0.6in]{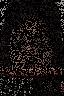}
  \label{fig:current_obsv}
  }\hfil
  \subfigure[]{
  \includegraphics[width=0.1\linewidth,height=0.6in]{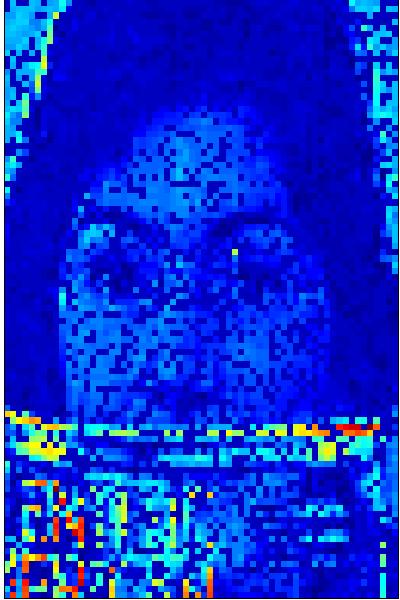}
  \label{fig:possibility}
  }\hfil
  \subfigure[]{
  \includegraphics[width=0.1\linewidth,height=0.6in]{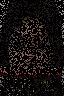}
  \label{fig:next_obsv}
  }\hfil
  \caption{Illustration of the online update of $\Omega$ between two consecutive frames. (a) The observed pixels of the current target. (b) The possibilities of the pixels to be indexed by new $\Omega$. The cooler pixel indicates the larger value. (c) The observed pixels of the next target, which are obtained according to the possibilities in (b).}
  \label{fig:update}
\end{figure}

Fig. \ref{fig:update} illustrates the online update strategy of $\Omega$ between two consecutive frames. It can be seen from Fig. \ref{fig:possibility} that the pixels from the distractive object (the magazine) have higher possibilities to be excluded (\ie, not indexed in $\Omega$) in the next frame. From Fig. \ref{fig:next_obsv}, it is evident that the pixels belonging to the distractive object are reduced in the local observations of the target in the next frame. In our experiments, similar to the work \cite{Mei2011}, we use ten target templates.

\section{Experimental Evaluations}
Our tracker is implemented in MATLAB on a PC with an Intel Core 2.8GHz processor. The average running speed is one frame per second. The colorful pixels in each frame are converted to gray scale and normalized to $\left[0,1\right]$. The corresponding regions of the candidates and the target templates are normalized to the size of $20\times20$ pixels, and 70\% pixels are observed for the candidates.

We compared our tracker with respect to 14 popular state-of-the-art trackers, including subspace methods, local methods, and other state-of-the-art methods. Their parameters were set to the values recommended by respective authors. In order to demonstrate the effectiveness of different algorithms in various challenging situations, we collect the most popular 20 video sequences for the comparisons, which demonstrate various challenges, such as heavy occlusions, illumination variations, and background clutters, as shown in Fig. \ref{fig:tracking_results}. In each frame, the target region is manually labeled using a bounding box as ground truth. Both the source codes and the video sequences can be publicly downloaded from the respective websites of the authors.

\subsection{Qualitative Evaluations}
\begin{figure*}[t]
\begin{center}
  \includegraphics[width=0.15\linewidth,height=0.4in]{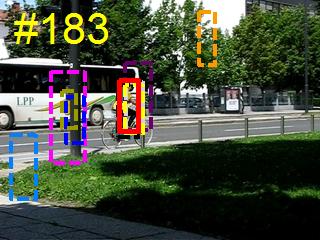}
  \includegraphics[width=0.15\linewidth,height=0.4in]{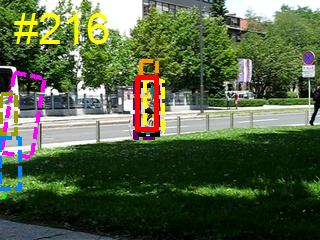}
  \includegraphics[width=0.15\linewidth,height=0.4in]{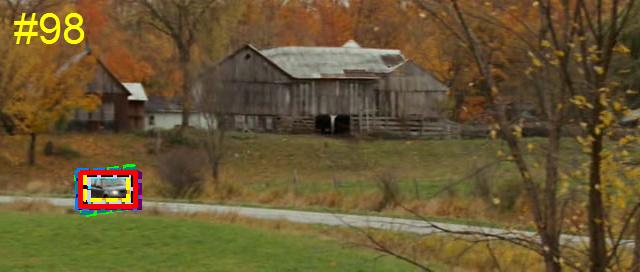}
  \includegraphics[width=0.15\linewidth,height=0.4in]{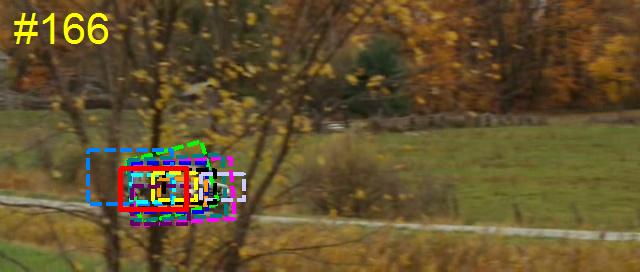}
  \includegraphics[width=0.15\linewidth,height=0.4in]{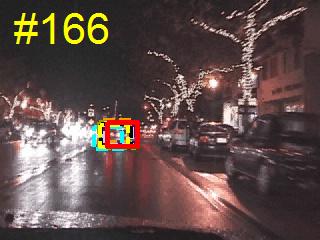}
  \includegraphics[width=0.15\linewidth,height=0.4in]{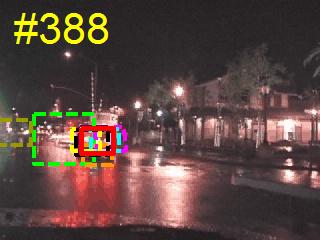}
  \includegraphics[width=0.15\linewidth,height=0.4in]{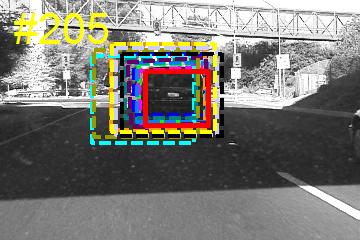}
  \includegraphics[width=0.15\linewidth,height=0.4in]{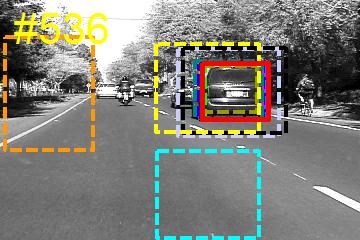}
  \includegraphics[width=0.15\linewidth,height=0.4in]{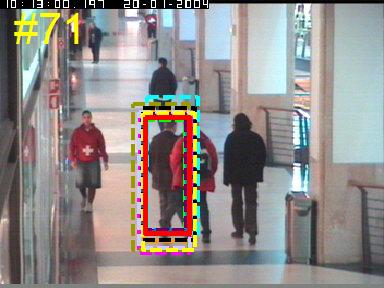}
  \includegraphics[width=0.15\linewidth,height=0.4in]{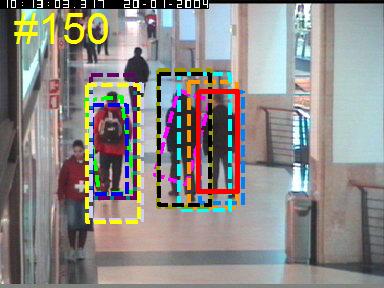}
  \includegraphics[width=0.15\linewidth,height=0.4in]{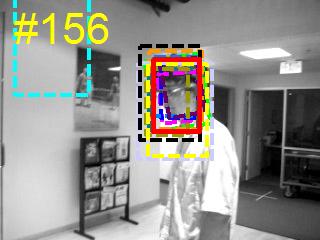}
  \includegraphics[width=0.15\linewidth,height=0.4in]{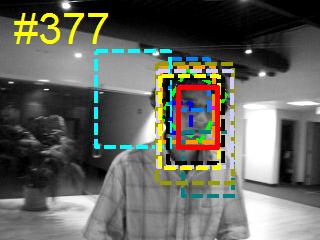}
  \includegraphics[width=0.15\linewidth,height=0.4in]{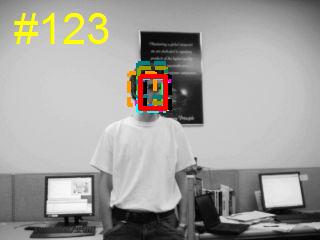}
  \includegraphics[width=0.15\linewidth,height=0.4in]{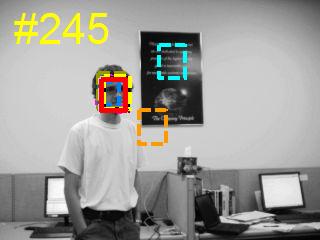}
  \includegraphics[width=0.15\linewidth,height=0.4in]{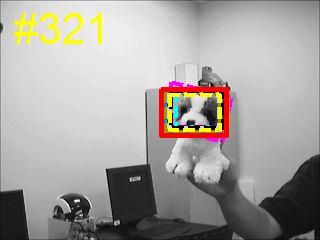}
  \includegraphics[width=0.15\linewidth,height=0.4in]{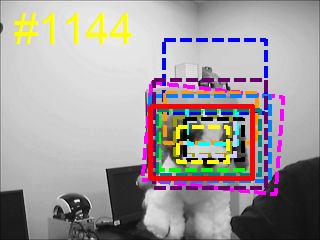}
  \includegraphics[width=0.15\linewidth,height=0.4in]{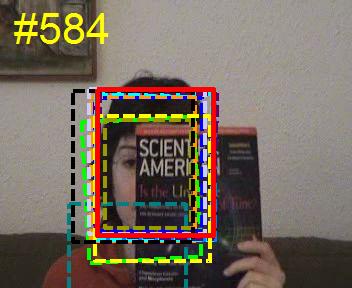}
  \includegraphics[width=0.15\linewidth,height=0.4in]{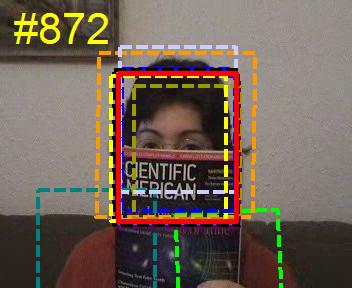}
  \includegraphics[width=0.15\linewidth,height=0.4in]{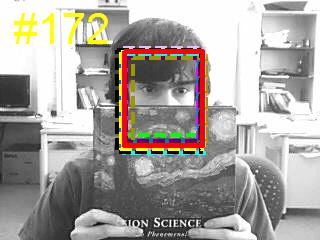}
  \includegraphics[width=0.15\linewidth,height=0.4in]{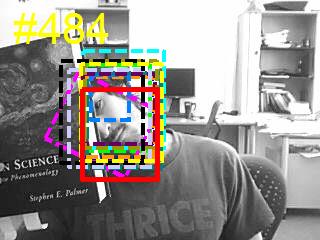}
  \includegraphics[width=0.15\linewidth,height=0.4in]{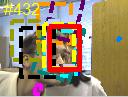}
  \includegraphics[width=0.15\linewidth,height=0.4in]{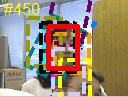}  \includegraphics[width=0.15\linewidth,height=0.4in]{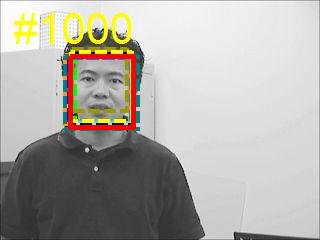}
  \includegraphics[width=0.15\linewidth,height=0.4in]{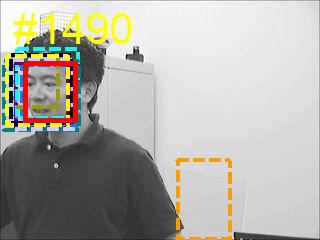}
  \includegraphics[width=0.15\linewidth,height=0.4in]{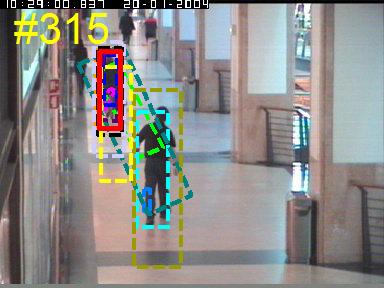}
  \includegraphics[width=0.15\linewidth,height=0.4in]{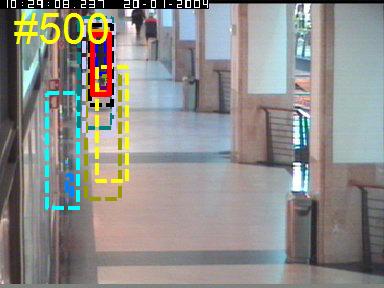}
  \includegraphics[width=0.15\linewidth,height=0.4in]{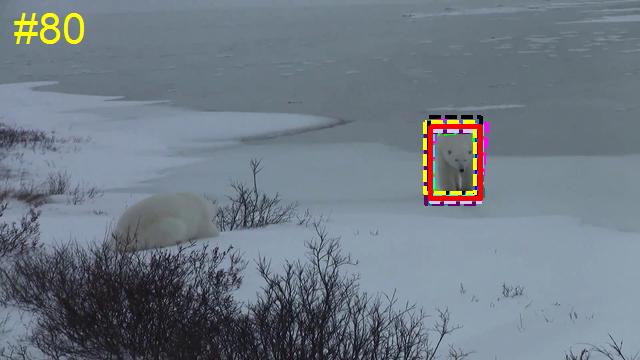}
  \includegraphics[width=0.15\linewidth,height=0.4in]{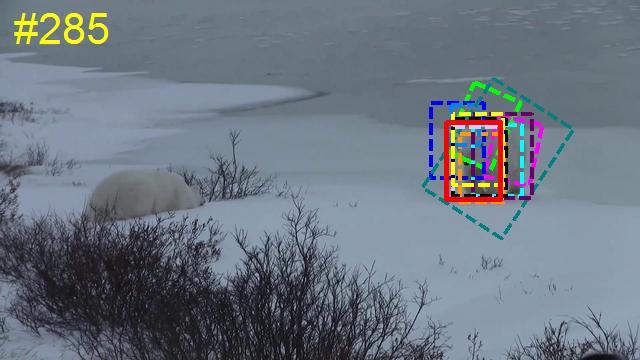}
  \includegraphics[width=0.15\linewidth,height=0.4in]{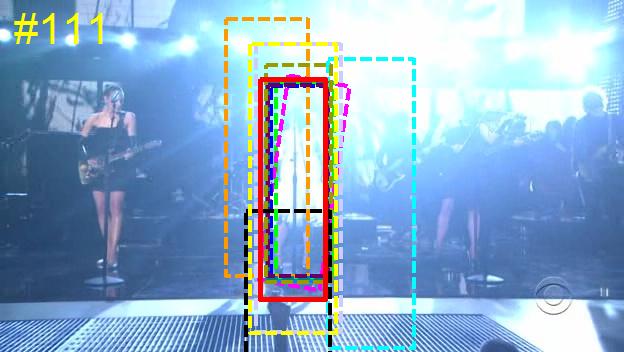}
  \includegraphics[width=0.15\linewidth,height=0.4in]{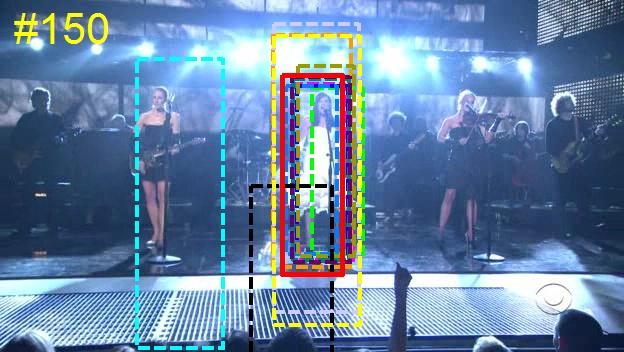}
  \includegraphics[width=0.15\linewidth,height=0.4in]{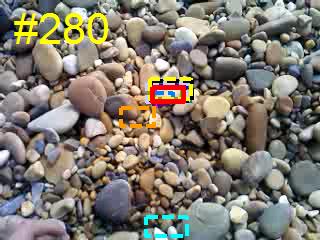}
  \includegraphics[width=0.15\linewidth,height=0.4in]{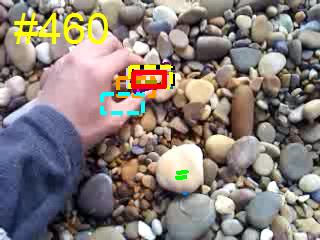}
  \includegraphics[width=0.15\linewidth,height=0.4in]{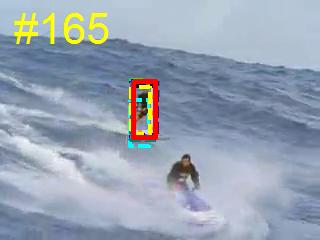}
  \includegraphics[width=0.15\linewidth,height=0.4in]{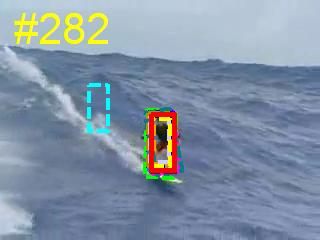}
  \includegraphics[width=0.15\linewidth,height=0.4in]{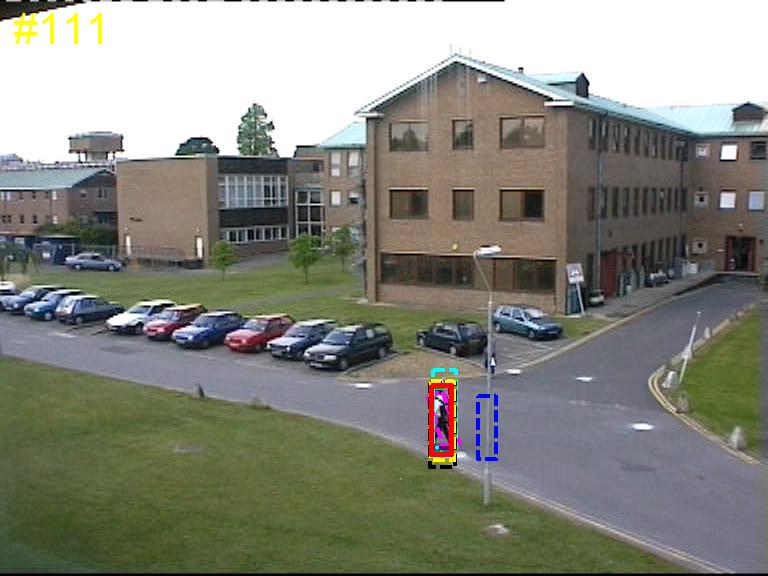}
  \includegraphics[width=0.15\linewidth,height=0.4in]{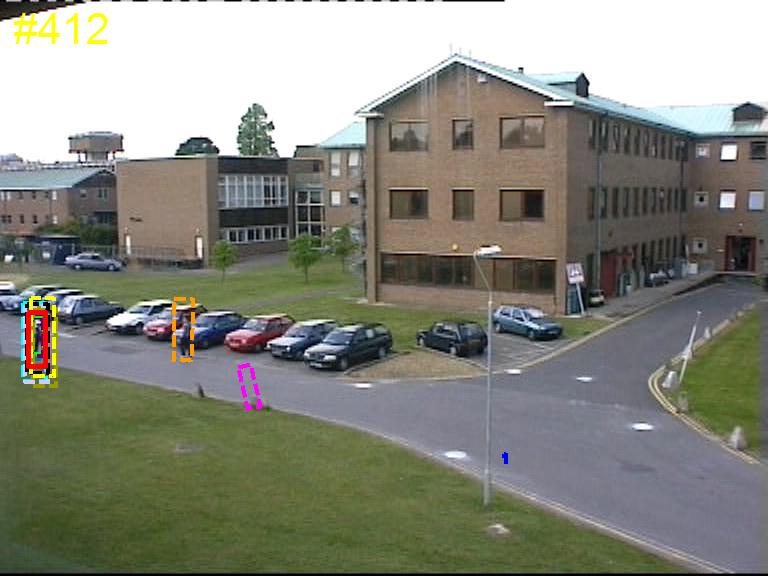}
  \includegraphics[width=0.15\linewidth,height=0.4in]{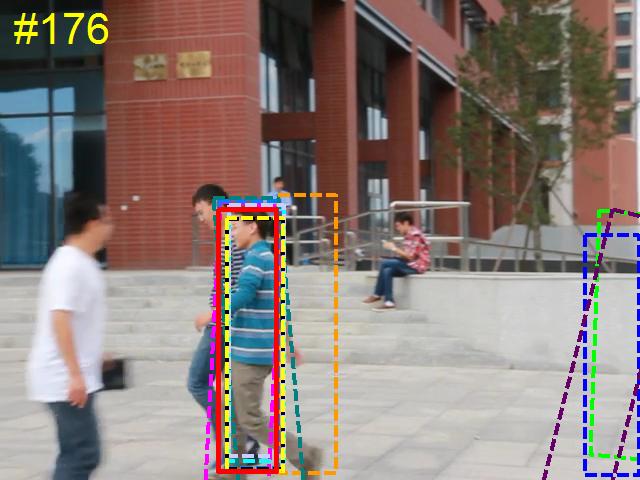}
  \includegraphics[width=0.15\linewidth,height=0.4in]{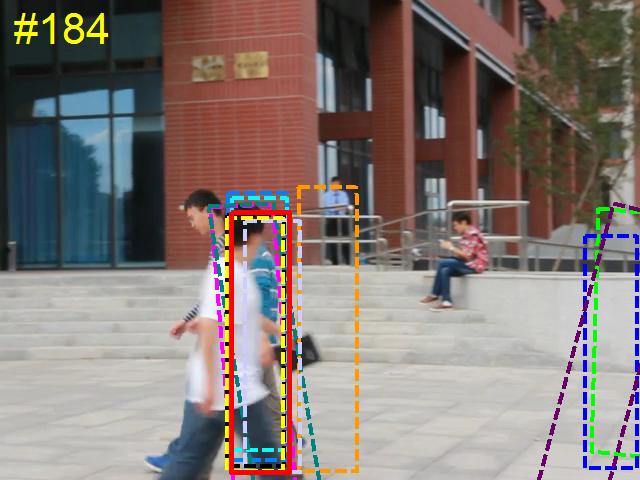}
  \includegraphics[width=0.15\linewidth,height=0.4in]{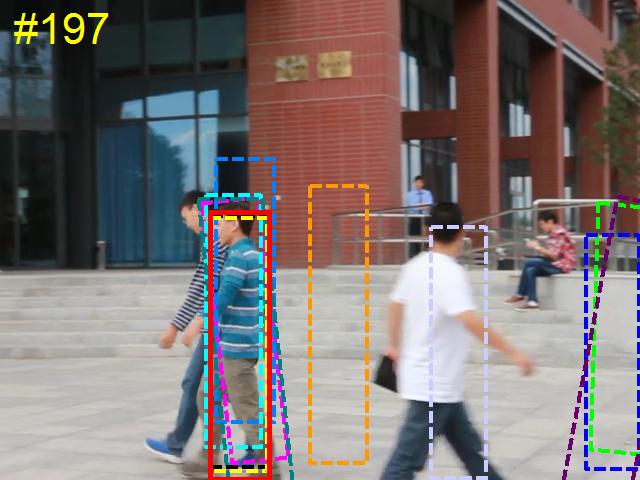}
  \includegraphics[width=0.15\linewidth,height=0.4in]{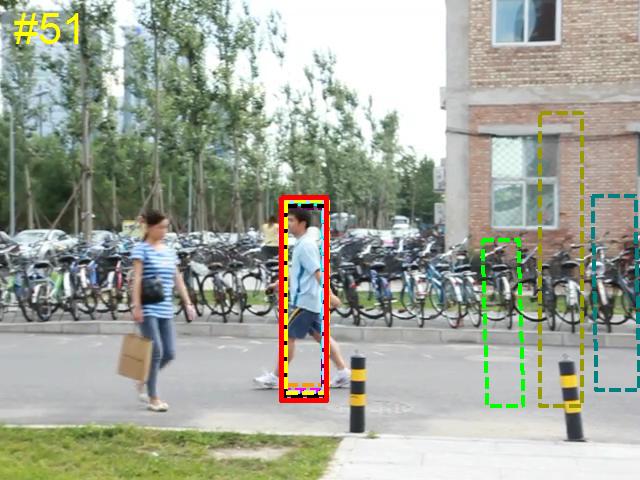}
  \includegraphics[width=0.15\linewidth,height=0.4in]{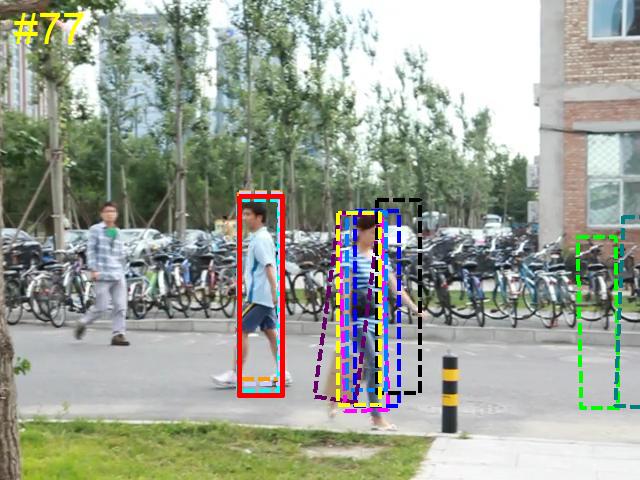}
  \includegraphics[width=0.15\linewidth,height=0.4in]{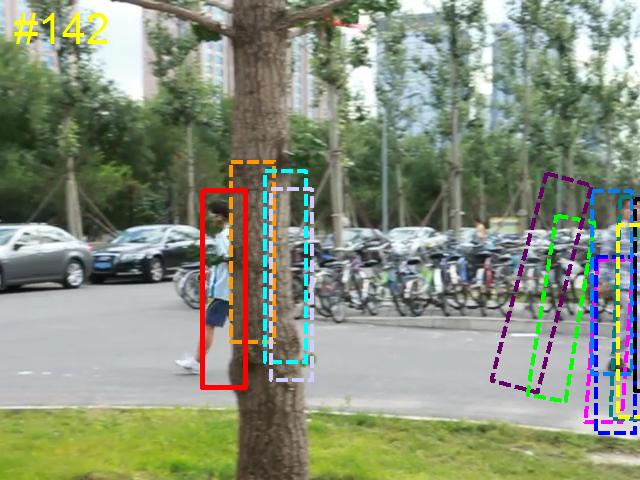}
  \includegraphics[width=0.94\linewidth]{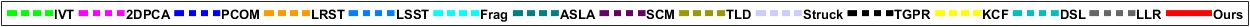}
\end{center}
   \caption{Tracking results on representative frames of the 20 video sequences.}
\label{fig:tracking_results}
\end{figure*}

Fig. \ref{fig:tracking_results} shows the tracking results obtained by our tracker and the 14 competing trackers on the representative frames of the 20 video sequences.

In the case of illumination changes, \eg, on the video sequences \emph{car4}, \emph{david} and \emph{singer1}, our tracker achieves better results, owing to the assumption of low-dimensional subspace on the previously obtained targets, which makes matrix completion work successfully. As a demonstration of the effectiveness of the subspace assumption, a very good tracking result is obtained by the proposed tracker in the case of pose changes, \eg, on the video sequences \emph{bicycle}, \emph{polarbear}, \emph{surfing} and \emph{walking}. In the presence of occlusions, \eg, on the video sequences \emph{bicycle}, \emph{caviar3}, \emph{faceocc}, \emph{oneleaveshop}, \emph{thusl} and \emph{thusy}, our tracker also achieves better or competitive tracking performance. This is attributed to: 1) the local observations and the online update strategy leverage the pixels that are not located on the distractive objects; and 2) the matrix completion leads to a high estimation accuracy.

\subsection{Quantitative Evaluations}
\begin{figure}[t]
  \centering
  \includegraphics[width=0.3\linewidth]{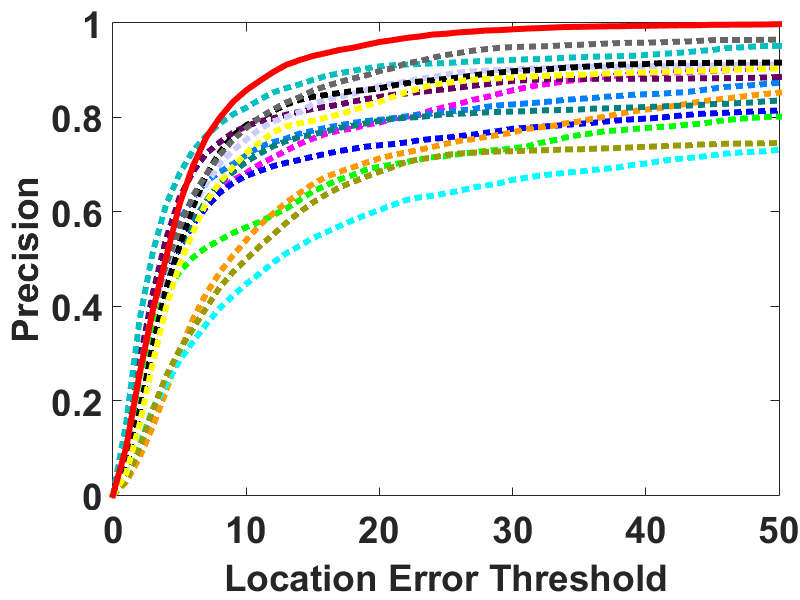}
  \hfil
  \includegraphics[width=0.3\linewidth]{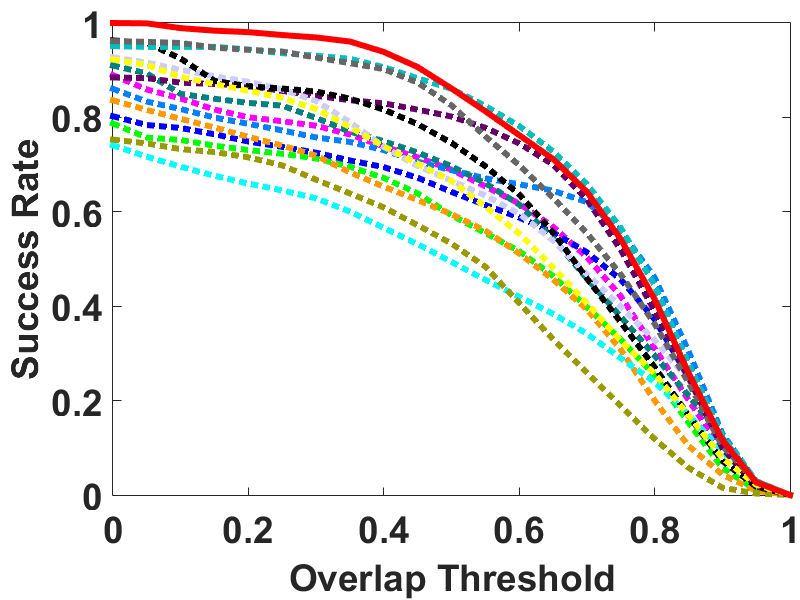}
  \includegraphics[width=0.9\linewidth]{imgs/legend}
  \caption{Tracking performance of the proposed and the 14 competing trackers on the 20 video sequences in terms of precision (left) and success rate (right).}
  \label{fig:perf}
\end{figure}

Four criteria are used to quantitatively evaluate the performance of different trackers: tracking location error (TLE), precision, overlap rates (OR), and success rate (SR). The TLE is computed from the difference between the centers of the tracking and the ground truth bounding boxes. The precision in defined as the percentage of frames where the TLE are less than a threshold $\delta$. The OR is computed by $\frac{A_T\cap A_G}{A_T\cup A_G}$, where $A_T$ and $A_G$ denote the areas of the bounding boxes of the tracking result and the ground truth, respectively. The SR is defined as the percentage of frames where the OR are greater than a threshold $\rho$. Table \ref{tab:results} shows the average TLE, precision (for $\delta=20$), OR, and SR (for $rho=0.5$) of ours and the 14 competing trackers on the 20 video sequences, respectively. Fig. \ref{fig:perf} plots the precision and the success rate of our tracker and its 14 counterparts on the 20 video sequences. From the quantitative evaluations on the 20 video sequences, it is evident that the proposed tracker outperforms its 14 counterparts in terms of all four criteria.
\setlength{\tabcolsep}{2.5pt}
\begin{table*}[t]
  \renewcommand{\arraystretch}{0.6}
  \caption{Tracking performance of the proposed and the competing trackers on the 20 video sequences. The best results are shown in bold-face font.}
  \label{tab:results}
  \centering
  \scalebox{0.7}[0.7]{
  \begin{tabular}{|c|c|c|c|c|c|c|c|c|c|c|c|c|c|c|c|}
    \hline
    tracker & ~Ours~ & IVT & 2DPCA & PCOM & LRST & LSST & Frag & ASLA & SCM & TLD & Struck & TGPR & KCF & DSL & LLR \\
    ~ & ~ & \cite{Ross2007} & \cite{Wang2012} & \cite{Wang2014} & \cite{Zhang2012b} & \cite{Wang2013} & \cite{Adam2006} & \cite{Jia2012} & \cite{Zhong2012} & \cite{Kalal2012} & \cite{Hare2011} & \cite{Gao2014} & \cite{Henriques2015} & \cite{Sui2015c} & \cite{Sui2016} \\
    \hline\hline
    \emph{tracking error} & \textbf{5.8} & 46.4 & 23.8 & 48.6 & 25.9 & 28.6 & 34.7 & 30.9 & 29.9 & 30.1 & 20.7 & 23.2 & 22.7 & 15.2 & 9.9 \\
    \emph{precision} & \textbf{0.96} & 0.70 & 0.79 & 0.74 & 0.71 & 0.79 & 0.60 & 0.79 & 0.84 & 0.68 & 0.87 & 0.86 & 0.83 & 0.91 & 0.90 \\
    \emph{success rate} & \textbf{0.86} & 0.59 & 0.69 & 0.64 & 0.59 & 0.69 & 0.49 & 0.70 & 0.80 & 0.53 & 0.67 & 0.75 & 0.67 & 0.85 & 0.83 \\
    \emph{overlap rate} & \textbf{0.72} & 0.51 & 0.58 & 0.56 & 0.51 & 0.50 & 0.47 & 0.59 & 0.65 & 0.50 & 0.60 & 0.61 & 0.57 & 0.70 & 0.68 \\
    \hline
  \end{tabular}
  }
\end{table*}

\subsection{Analysis of the Tracking Model}
The impressive performance of the proposed tracker is attributed to the integration of the global subspace assumption and the local observations. This is further demonstrated below from an experimental perspective.

\begin{figure}[t]
  \begin{center}
  \subfigure[]{
  \label{fig:dim}
  \includegraphics[width=0.225\linewidth]{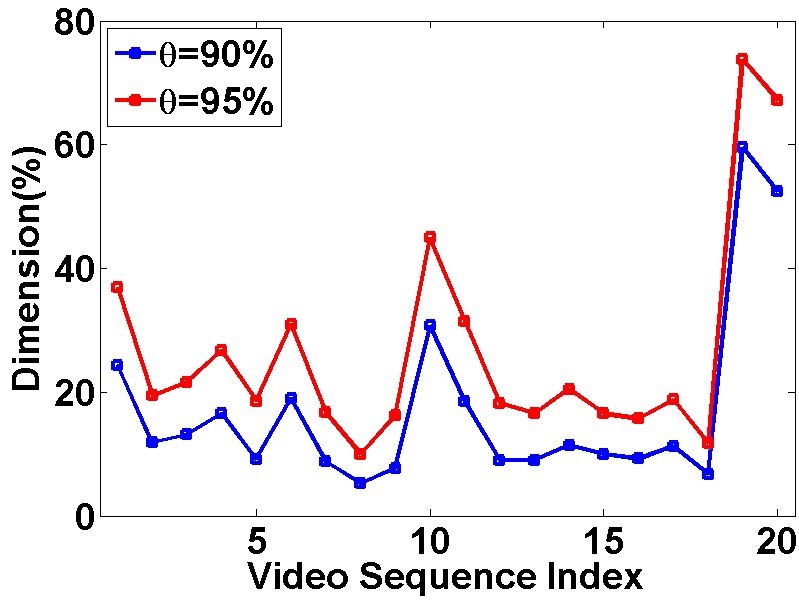}
  }
  \subfigure[]{
  \label{fig:local}
  \includegraphics[width=0.225\linewidth]{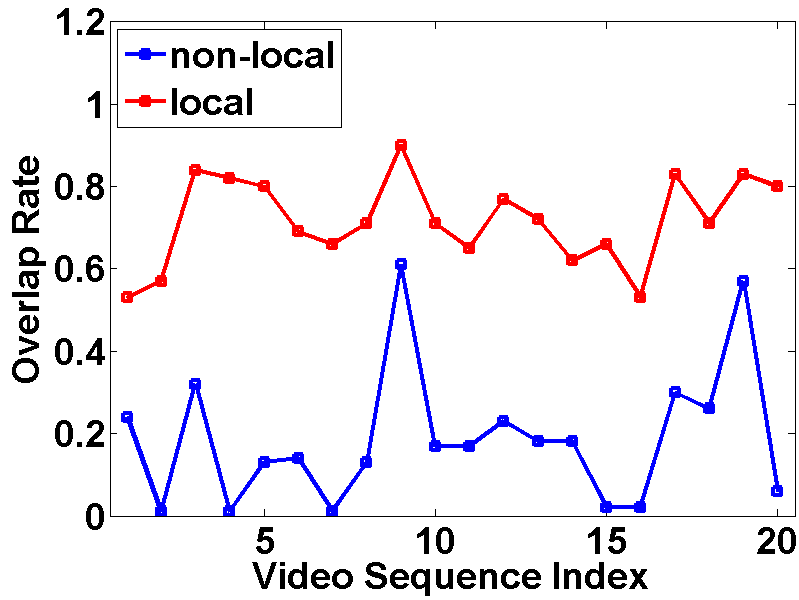}
  }
  \subfigure[]{
  \label{fig:TLE}
  \includegraphics[width=0.225\linewidth]{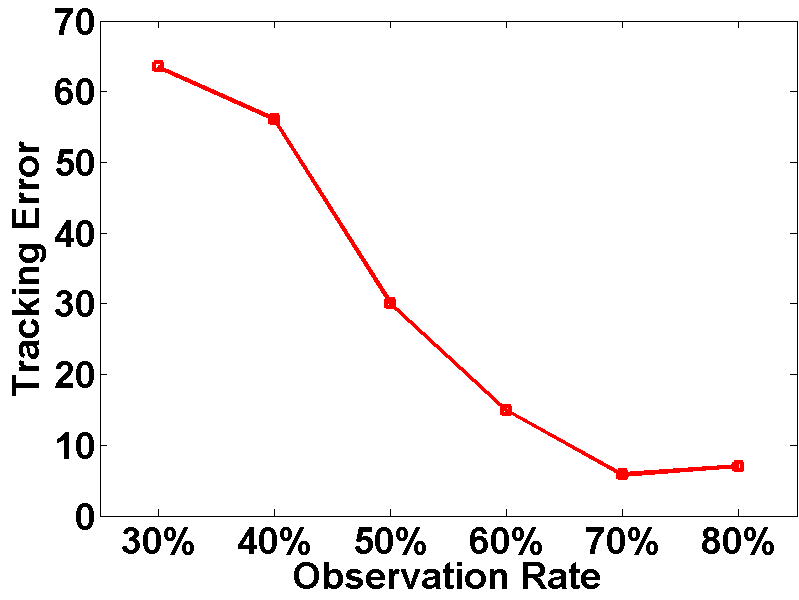}
  }
  \subfigure[]{
  \label{fig:OR}
  \includegraphics[width=0.225\linewidth]{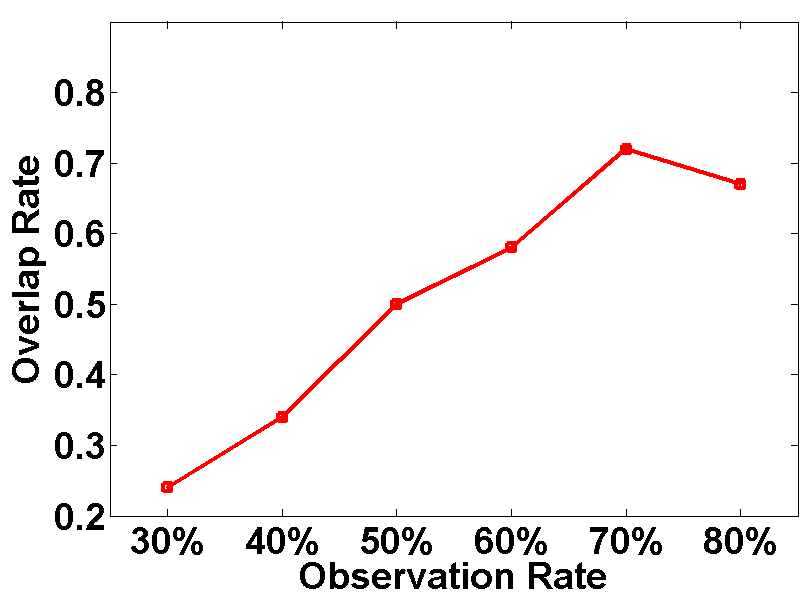}
  }
  \caption{Low dimension degrees (a) and overlap rates of the trackers with and without local observations (b) on the 20 experimental sequences. The order of the video sequences is identical to that showed in Fig. \ref{fig:tracking_results}. Tracking location errors (c) and overlap rates (d) on the 20 video sequences with respect to different local observation rates.}
  \end{center}
\end{figure}

First, we verify the assumption that the targets reside in a low-dimensional subspace. We collect all the targets from each of the 20 experimental video sequences according to their ground truths. Then, we analyze the corresponding target subspace by using singular value decomposition. If the target subspace is of low-dimensional, the metric of \emph{low dimension degree}, defined as the number of the non-zero singular values of the target matrix, whose sum is more than $\theta$ times the sum of all singular values, should be small for a large $\theta$. Fig. \ref{fig:dim} shows the results on the 20 video sequences with respect to $\theta=90\%$ and $\theta=95\%$, respectively. It can be seen that most of the low dimension degrees are located within the range of $\left[10\%,40\%\right]$, which indicates that the targets truly reside in a low-dimensional subspace.

\begin{figure}[t]
  \begin{center}
  \includegraphics[width=0.48\linewidth]{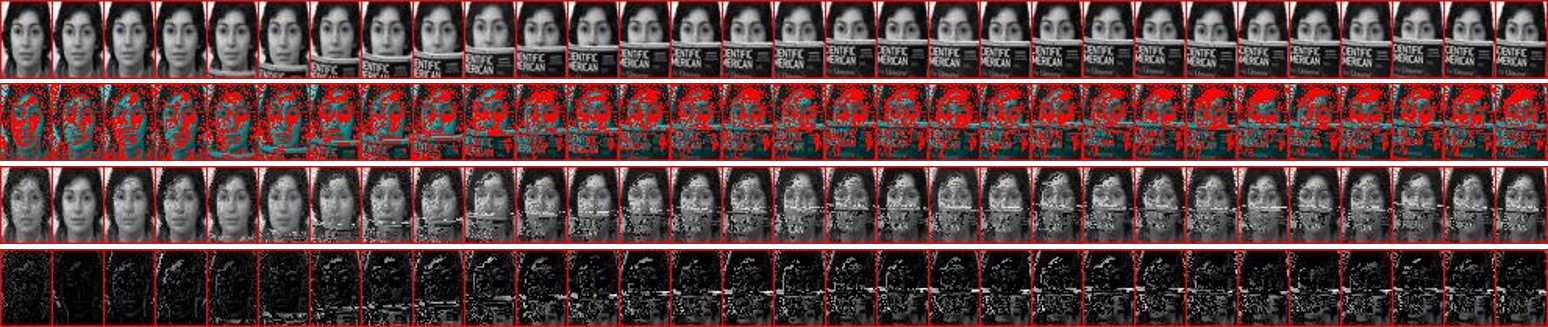}
  \includegraphics[width=0.48\linewidth]{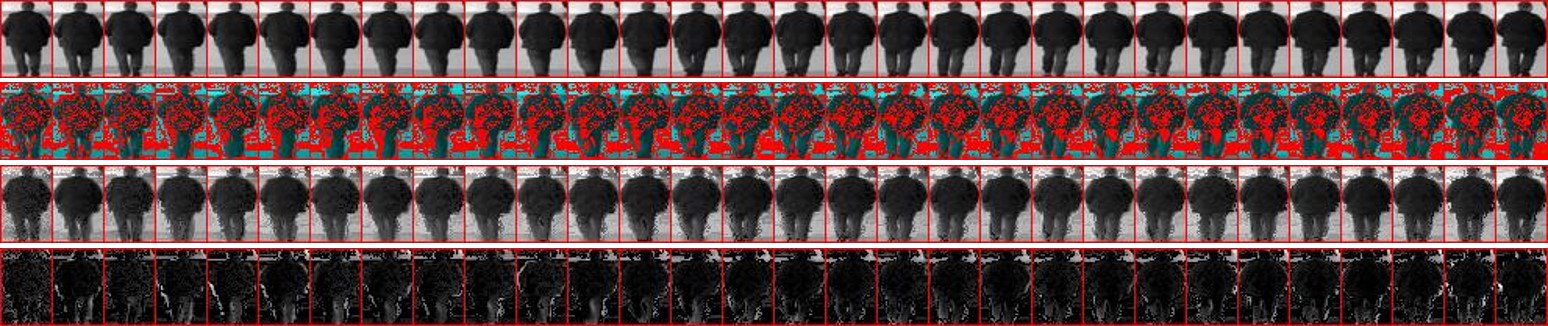}
  \caption{Visualization of the local observation. The 1st row show the temporally obtained targets. The red pixels in the 2nd row indicate the observed pixels. The estimated targets and the residual errors are shown in the 3rd and the 4th rows, respectively.}
  \label{fig:visualization}
  \end{center}
\end{figure}

Next, we demonstrate the effectiveness of the integration of the local observations with the global subspace assumption. We compare the tracking results of the two trackers with and without local observations on the 20 video sequences. The results are shown in Fig. \ref{fig:local}, from which we can see that the tracking performance is significantly improved by using the local observations. We also visualize the local observation in Fig. \ref{fig:visualization}. It can be seen that, in the case of occlusion, the book tends to be observed with a small possibility, and in the case of deformation, the body of the person and the static surrounding background are encouraged to be observed, while the deformations (on shoulders and legs) are rarely observed. These experiments indicate that the local observations provide strong priors for the estimator (matrix completion), leading to more accurate estimations.

Furthermore, we show how the local observations influence the tracking results. Since the number of the observed pixels is a critical factor for the tracker, we investigate the TLEs and ORs with respect to different numbers of the observed pixels, as shown in Fig. \ref{fig:OR}. It is evident that our tracker yields the best performance when about $70\%$ pixels are observed. Observing too few pixels may lead to an inaccuracy of the matrix completion, while observing too many pixels may result in heavy influence from distractive objects. As a result, we observe $70\%$ pixels in our experiments for each frame.

\section{Conclusion}
We have formulated tracking as a problem of target appearance estimation by exploiting the advantages of both the global and local tracking models. Extensive experiments have been conducted and the results have demonstrated that: 1) our tracking model, by integrating the global and the local methods, effectively alleviates tracking failures in various challenging situations; and 2) under the subspace assumption, the matrix completion provides an accurate estimation in target appearance for the target location. Both the qualitative and the quantitative evaluations have demonstrated that the proposed tracker outperforms most popular state-of-the-art trackers.

\section*{Acknowledgement}
The work is partly supported by the National Natural Science Foundation of China (NSFC) under grants 61273282, 61573351 and 61132007, and the joint fund of Civil Aviation Research by the National Natural Science Foundation of China (NSFC) and Civil Aviation Administration under grant U1533132.

\clearpage


\end{document}